\documentclass{ifacconf}

\usepackage{graphicx}      %
\usepackage{natbib}        %

\usepackage{subcaption} %
\usepackage{siunitx} %
\usepackage{amsmath} %
\usepackage{amssymb} %
\usepackage{multirow} %
\usepackage{xcolor} %

\usepackage{geometry}
\newtheorem{definition}{Definition}
\newtheorem{remark}{Remark}

\DeclareMathOperator{\diag}{diag}
\begin{document}
    \begin{frontmatter}

        \title{A Generalized Approach to Impedance Control Design for Robotic Minimally Invasive Surgery\thanksref{footnoteinfo}}

        \thanks[footnoteinfo]{D. Larby is partially supported by the EPSRC grant EP/T517847/1, and by CMR Surgical}

        \author[First]{Daniel Larby}
        \author[First]{Fulvio Forni}

        \address[First]{ Department of Engineering, University of Cambridge, CB2 1PZ UK (e-mail: dl564@cam.ac.uk, f.forni@eng.cam.ac.uk).}

        \begin{abstract}                %
            Energy based control methods are at the core of modern robotic control algorithms.
            In this paper we present a general approach to virtual model/mechanism control, which is a powerful design tool to create energy based controllers.
            We present two novel virtual-mechanisms designed for robotic minimally invasive surgery, which control the position of a surgical instrument while passing through an incision.
            To these virtual mechanisms we apply the parameter tuning method of \cite{Larby2022}, which optimizes for local performance while ensuring global stability.
        \end{abstract}

        \begin{keyword}
            Passivity-based control; Lagrangian and Hamiltonian systems; Robotic Surgery\end{keyword}

    \end{frontmatter}

    \section{Introduction}

    In Robotic Minimally Invasive Surgery the robot enters the patient through a trocar---a surgical device used to create and maintain a portal into the patient's body for the insertion of surgical instruments.
    The robot's end-effector follows the surgeon's motion, captured by the surgeon console.
    Usually the surgeon orchestrates at least 2 instruments and an endoscope (a minimally invasive camera) simultaneously \cite{Klodmann2021, Thai2020a, Chen2020}.

    Different types of robotic minimally invasive surgery pose different challenges.
    Video Assisted Thoracoscopic Surgery (VATS), which takes place in the chest, must intrude upon the ribcage.
    In arthroscopy (keyhole surgery of a joint such as a hip or knee) the workspace is extremely limited by the anatomy of the joints, and the trocar much closer to the site of the operation than during, for example, abdominal laparoscopy.

    Typically, the surgeon exclusively controls the motion of the end-effector and the robot ensures that the instrument passes through the trocar, either by a suitable mechanism or by coordinated control of the joints, \cite{Chen2020}. In the latter case the position the trocar, the \emph{Remote Center of Motion} (RCM), is typically recorded at the beginning of the procedure, denoted  and used to constrain the motion of the instrument.

    In this paper we advocate that passing through the RCM is an oversimplified representation of the trocar task for some procedures.
    For example in thoracoscopic surgery, the chest wall may restrict movement, ignorance of which can result in ``localized trauma at the incision site or compression of the intercostal nerve'' \cite{Trevis2020}.
    Conversely, sometimes we may wish to relax the RCM constraint to improve mobility of the end-effector, for example by expanding the RCM constraint from a point to a hoop.
    This could be beneficial for procedures such as trans-oral surgery, arthroscopy, small volume-laparoscopy.

    \begin{figure}[htbp]
        \centering
        \includegraphics{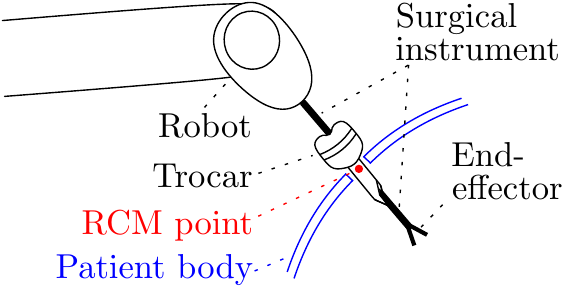}
        \caption{Section view of robotic minimally invasive laparoscopy, highlighting the trocar and Remote Centre of Motion (RCM).}
    \end{figure}

    In this paper we propose a method for the design of robot controllers for a general trocar task,
    which takes advantage of a more complex representation of the constraints that limit the motion of the instrument.
    The controllers are constructed to emulate a virtual mechanism.
    This technique is extremely flexible, as the intuition of the virtual mechanism guides the design.
    Furthermore, the emulation of a virtual mechanism guarantees that the control action is passive,
    with the performance determined by the choice of the control parameters.

    The design can be split into two tasks: designing the mechanism's structure,
    and deciding the mechanism's parameters (e.g. spring stiffnesses, damping coefficients).
    Both steps are important  but there is a lack of methods for tuning parameters automatically, therefore we also present an approach to tuning parameters for a certain class of mechanisms, to optimize robots performance around chosen poses while maintaining the passivity promised by the virtual mechanism approach. The structure is designed to be amenable a
    performance led parameter synthesis, and desired relative performances of the port and end-effector are used in the synthesis.

    Section \ref{sec:ControllerStructure}, presents the general approach to designing a controller structure as a virtual mechanism and implementing it in control.
    Section \ref{sec:FundamentalComponents} introduces the fundamental components of such a mechanism.
    In Section \ref{sec:VirtualMechanismsForSurgery} we apply this approach to robotic minimally invasive surgery.
    In Section \ref{sec:Tuning} we demonstrate our parameter tuning technique, and in Section \ref{sec:Results} we present results in simulation.

    \section{Passive structures for control}  \label{sec:ControllerStructure}

    \subsection{Control as interconnection with virtual mechanisms}
\begin{figure}[htbp]
        \centering
        \begin{subfigure}{0.3\columnwidth}
            \centering
            \vspace{5mm}
            \includegraphics[scale=0.8]{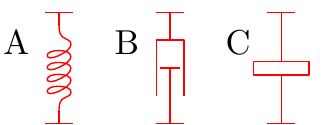}
            \vspace{1.cm}
            \caption{}
        \end{subfigure}%
        \hfill
        \begin{subfigure}{0.6\columnwidth}
            \centering
            \includegraphics{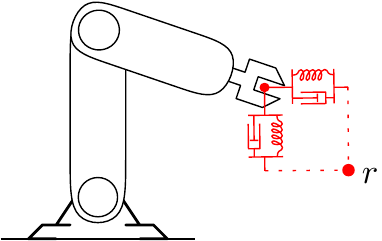}
            \caption{}
            \label{fig:TwoLinkCartesian}
        \end{subfigure}
        \caption{a) The three fundamental components. A: a spring, B: a damper, C: an inerter.
            b) Illustration of Cartesian Impedance controller as a virtual mechanism.}
    \end{figure}

    The control design approach we suggest can be described as follows.
    We envision a mechanism to achieve the robot's task, and implement it as a \textit{virtual mechanism} driving the controller action on the robot.
    From this perspective jointspace PD control can be viewed as a mechanism of rotational spring-dampers at the robot's joints, and Cartesian control can be viewed as a mechanism of Cartesian springs and dampers attached between the end-effector and a reference position, as shown in Figure \ref{fig:TwoLinkCartesian}.
    Following the virtual mechanism approach we can go beyond simple combinations of springs and dampers, using them in creative ways to achieve complex tasks.

    We take inspiration from \textit{Virtual Model Control} as described in \cite{Pratt2001}. Previous work on virtual model control seems focussed on mobile robotics such as walkers or quadrupeds \cite{Pratt2001, Desai2014, Chen2020a}.
    In contrast, for surgery (and for manipulation , in general), we assume that the robot is rigidly attached to the world.

    This approach has many benefits.
    The virtual mechanism is an Euler-Lagrange system---like the robot---and thus when connected in feedback with the robot, achieves closed-loop passivity.
    Fundamentally, designing a virtual mechanism and emulating it with a robot controller is a form of energy-shaping and damping injection (\cite{Ortega1998}), where the structure of the virtual mechanism provides a set of basis-functions/primitives to use.
    Therefore, the closed-loop is passive at any port, which implies that local minima of its potential energy are all stable (and asymptotically stable when damping is present).
    These features are preserved even under interaction with the environment, so long as the environment is passive.
    The controller implementation is simple: no inverse kinematics required.
    It can handle multiple simultaneous objectives---such as RCM and end-effector control.
    Finally, the virtual mechanism is an excellent tool to communicate with practitioners of many disciplines and produces controllers that are inherently more interpretable by users.
    This could lead to better communication with the control engineer, and easier/more effective debugging.

    To design such a mechanism, we begin with a library of ideal components, and interconnect them with the robot at various ports. %
    These components are ideal dampers, springs and inerters, which we will use to construct our controller.
    They correspond to a proportional, integral, and derivative relationship between effort and flow and are analogous with the linear, passive electrical components: the ideal resistor, capacitor, and inductor.

    \subsection{Realization of Virtual Mechanisms}

    The interconnection of our virtual mechanism with the robot is achieved by a transformation from the joints of the robot to various `operation-space' coordinates.

    The simplest example of an operation-space coordinate is the velocity of a point which is fixed relative to a link of the robot, such as the velocity of the end-effector.
    A port for this operation-space coordinate has the flow/effort pair of $\dot z$ and $F$, where $\dot z$ is the velocity of the end-effector in the world-frame, and $F$ is a force vector in the same (world) frame as the velocity, such that $\dot z ^T F$ is the power flow into the system.

    We take a broad interpretation of operation space variables: an operation space velocity does not need to simply be the world-frame velocity of a point on the robot, but could for example be the link-frame velocity of a fixed point in the world, or the relative velocity of two different points on a robot.
    Spring-dampers on these coordinates would be aligned with link-axes rather the world's xyz axes, or allow us to attach a spring between two points.
    The choice of operation space coordinate allows us to choose `where' a component is attached, and the operation space Jacobian is the formal tool that allows us to convert efforts and flows between operation space and joint-space.

    Consider a robot with dynamic equation:
    \begin{equation}
        \label{eq:robotDynamics}
        M(q)\ddot q + C(q, \dot q)\dot q + g(q) = u,
    \end{equation}
    with inertia and Coriolis matrices, $M$ and $C$, and gravity  term $g(q)$.
    Given an operation space coordinate $z = h(q)$, its velocity is $\dot z = \frac{\partial h(q)}{\partial q} \dot q = J(q) \dot q$.
    If we wish to emulate the effect of force $F$ at $z$, generated by a generic ideal component and realised with the control input $u$ at the joints, we must apply $u = J(q)^T F$ (Section 9.4.1 of \cite{Spong1989}).
    Given this fact, we can apply an arbitrary force across any operation-space coordinate which we will use to implement springs, dampers and inerters.

    \begin{figure}[htbp]
        \centering
        \includegraphics[width=.99\columnwidth]{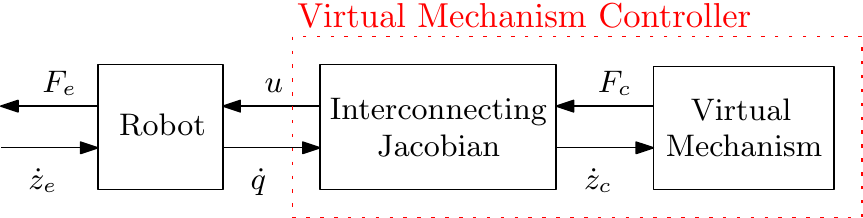}
        \caption{Block diagram of the virtual mechanism controller, and passive interconnections}
        \label{fig:PortHamiltonian}
    \end{figure}

    \begin{definition}
        The virtual mechanism controller for robot \eqref{eq:robotDynamics} acts on the generalized force of the robot, $u$, via feedback from the generalized coordinates, $q$ and $\dot q$. Its action is generated by emulating a virtual mechanism connected to the robot at operation-space ports $(F, \dot z)$. Its action is realized through the operation space Jacobian that acts as a lossless transformer\footnote{Modulated by $q$, \cite{VanDerSchaft2020}.} between the control ports $(u, \dot q)$ and the operation space ports $(F, \dot z)$. \hfill $\lrcorner$
    \end{definition}

    The virtual controller will feature energy storages such as springs, inerters, masses or inertias; energy dissipation such as dampers; and energy routing elements such as rigid-body joints, levers, and mechanical transmissions.
    As a virtual mechanism it will feature classical passivity properties at co-located force/velocity or torque/angular velocity ports.

    \begin{thm}
        \label{the:stable}
        The closed-loop system featuring a robot and passive virtual mechanism controller, represented in Figure \ref{fig:PortHamiltonian}, is passive at any co-located generalized force / generalized velocity port $(F_e, \dot z_e)$.
    \end{thm}
    \begin{pf}
        The robot and virtual mechanism controller are passive at their ports.
        The interconnection Jacobian acts as a lossless transformer: the power flow in equals the power flow out $u^T \dot q = F_c^T J(q) \dot q =F_c^T \dot z_c$.
        Thus, their interconnection is passive with respect to $(F_e, \dot z_e)$. \hfill $\square$
    \end{pf}

    \subsection{Fundamental Components} \label{sec:FundamentalComponents}

    We will briefly describe a set of basic components that can be used to construct virtual mechanisms.

    \emph{Virtual springs and potential energy shaping ---}
    Virtual springs are possibly the most important component of a practical virtual mechanism.
    They drive the robot to the ``correct'' position, determining the equilibria of the system.

    The characteristic of a nonlinear spring $F = f_s(z)$, where $z$ refers to the spring's extension and $k>0$ the spring stiffness, shapes the energy of the system.
    Different effects such as deadzones or saturations can thus be achieved.
    The only proviso is that the energy stored in the spring $E(z)$ is bounded below, that is,
    $\frac{\partial E(z)}{\partial z} = f_s(z)$ and there exists a finite constant $C$ such that $E(z) > C$ for all $z$.
    Otherwise, an infinite amount of energy could be extracted from the spring.
    $E(z)$ reduces to $k z^2$ for linear springs $f_s(z)=kz$.

    \emph{Dampers and damping injection ---}
    Dampers are the second vital component of a passive robot controller.
    Robots, as mechanical systems, have some inherent damping from friction, but the injection of additional damping is needed to shape performance, as damping affects \textit{how} the system approaches equilibrium.

    An ideal nonlinear damper $F = f_d(z, \dot z, t)$ must satisfy the instantaneous passivity condition $f_d(\cdot)\dot z \ge 0$; a damper can only remove energy from the system. $z$ and $\dot{z}$ refer to displacement and velocity across the damper. A simple parametrization is given by $f_d(z, \dot z, t) = c(\cdot) \dot z$ where $c(\cdot) \geq 0$. The simplest form of a linear damper reads $f_d(z, \dot z, t) = c \dot z$, which is passive so long as $c \geq 0$.

    \emph{Inerters and kinetic energy shaping ---}
    An ideal, linear inerter, \cite{Smith2020}, produces a force proportional to an acceleration: $F = m \ddot z$.
    $\ddot z$ is the acceleration between the two ends of the inerter, and $m$ is the inertance, which has units of \SI{}{kg}.
    An inerter generalises the concept of mass.
    The dynamics of a point-mass m can be recreated by combining three inerters at a point, one along each Cartesian axis, having inertance m and their other terminal grounded.
    Inerters are rarely used in passivity-based control of robots; control engineers usually opt for a Proportional/Derivative approach\footnote{Proportional/Derivative from position, i.e. Proportional/Integral from velocity}.
    This might be due to difficulty in sensing/estimating acceleration, or due to difficulty in actuation: acceleration has more high frequency content than position or velocity signals, so a wider bandwidth is required.

    Inerters are useful when paired with dynamic extensions: additional revolute/prismatic joints, that form part of the virtual mechanism.
    A virtual joint followed by a massless virtual link would experience infinite acceleration.
    A virtual link with some virtual mass causes practical difficulties: if the acceleration of the virtual mass depends not just upon the virtual joints acceleration, but also the robot's joint accelerations, then the joint accelerations must be measured or estimated to implement the inertial forces related to the virtual mass.
    Instead, by placing a virtual inerter directly across the virtual joint (as will be demonstrated in Section \ref{sec:PrismaticExtension}), we limit the acceleration of the virtual joint, which is no longer infinite, and we avoid having to measure or estimate of joint-accelerations.

    \section{Virtual Mechanisms for Minimally Invasive Surgery} \label{sec:VirtualMechanismsForSurgery}

    In this section we use the described approach to derive new structures for the problem of robotic minimally invasive surgery.
    We describe two virtual mechanisms/controller structures.
    Because we follow the principles outlined above our design will be passive and robust. The intuitive mechanical analogy acts as a baseline for further modifications to the control structure, to adapt it to different surgical challenges.

    We contribute the design of novel virtual mechanisms for the RMIS task, and the design of mechanisms amenable to the parameter synthesis technique from \cite{Larby2022}.
    Furthermore, we highlight the utility of the virtual inerter in a robotics application, other examples of which we are currently unaware.

    The first mechanism will be named as the ``Prismatic Extension,'' and the second as the ``Virtual Instrument.'' A feedforward gravity compensation of the robot is assumed for both mechanisms, which is compatible with passive energy shaping (\cite{Lachner2022}) as the gravitational potential energy of a revolute-jointed robot is bounded above and below.
    Both mechanisms are illustrated in 2D (for ease of exposition).
    Generalization to a 3D is achieved by the use of out-of-plane springs where applicable, not depicted.

    \subsection{Prismatic Extension} \label{sec:PrismaticExtension}

\begin{figure}[h]
        \centering
        \includegraphics{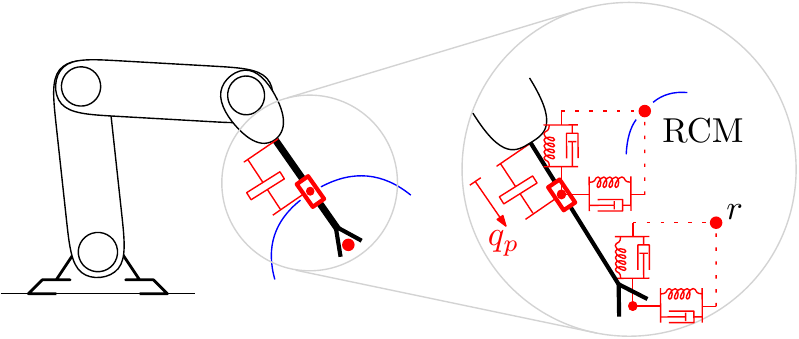}
        \caption{Illustration of the `Prismatic Extension' virtual mechanism.
            The red rectangle indicates a virtual prismatic joint that can slide along the instrument, with extension $q_p$ and a virtual inerter placed across the joint.
            The virtual spring-dampers constrain the slider to the remote centre of motion (RCM), and the end-effector to the surgeon reference, $r$.}
        \label{fig:PrismaticExtension}
    \end{figure}

    The virtual-mechanism of the controller is illustrated in Figure \ref{fig:PrismaticExtension} in red.
    Each component performs a specific function; we arrive at this structure by progressively assembling individual components with individual functions.

    The end-effector must follow the surgeon's command, which we achieve by attaching springs to the end-effector along Cartesian axes and connect them to the reference $r$.
    We wish for the instrument to pass through the trocar, but still allow sliding motion along the axis of the instrument.
    To enable the sliding motion we create a virtual-prismatic-joint affixed to the base of---and aligned with the axis of--the instrument, then will restrict the far end of the prismatic joint to the remote centre of motion (RCM), using Cartesian springs.

    To have a well-defined acceleration, the virtual prismatic joint  must experience some inertia.
    This is achieved by placing an inerter directly across the joint (as in Figure \ref{fig:PrismaticExtension})
    \[F_p = m_c \ddot q_p.\]
    The inerter defines the joint's acceleration-force characteristic, but doesn't affect any other joints, so $\ddot q_p$ depends only on the force across the virtual joint.

    The result is an extended system (robot + dynamic extension) with extended inertia matrix \[M_e(q_e) = \begin{bmatrix}
        M(q) & 0_{N\times1} \\
        0_{1\times N} & m_c
    \end{bmatrix},\]
    which is block diagonal; the robot's inertia is unaffected by the dynamic extension's inertia.
    Thus, it behaves similarly to a link with virtual mass/inertia, without the difficulty of inertia being reflected to other (non-virtual) joints.
    This leads to a simpler implementation. In contrast,  modifying the apparent inertia of the real robot joints would require joint-acceleration measurements/estimations, and higher-bandwidth control.

    The final addition the virtual mechanism is damping. We include 4 dampers co-located with the springs.
    While the placement of springs is done based on intuition from the task, the optimal placement of dampers is not so clear.
    We choose to co-locate the dampers with the springs to allow for algorithmic parameter selection in Section \ref{sec:Tuning}.

    \subsection{Virtual Instrument} \label{sec:VirtualInstrument}

    The second mechanism is constructed around the idea of a `virtual instrument': a dynamic extension consisting of a revolute joint fixed in the world at the RCM, followed by a prismatic joint, with the axis passing through the RCM. The virtual link is endowed with mass $m_i$ and inertia matrix $I_i$, so that the dynamic extensions inertia matrix $M_c(q_c)$ is never singular. By construction the axis of the virtual instrument always intersects the RCM.

    \begin{figure}[h]
        \centering
        \includegraphics{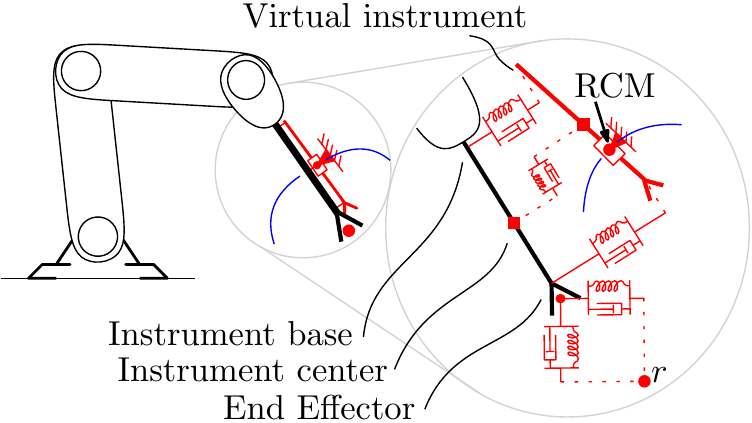}
        \caption{Illustration of the `Virtual Instrument' virtual mechanism.
            The virtual instrument (solid red line) always passes through the RCM by its construction, and is attached to the instrument by three spring-dampers.
            The virtual spring-dampers also constrain the end-effector to the surgeon reference.}
        \label{fig:VirtualInstrument}
    \end{figure}

    The surgical instrument is constrained to lie exactly on top of the virtual instrument by springs with a resting length of zero, as shown in Figure \ref{fig:VirtualInstrument}.
    With this non-rigid coupling the extended-system inertia is block diagonal
    \[M_e(q_e), = \begin{bmatrix}
        M(q) & 0_{N\times N_c} \\
        0_{1\times N_c} & M_c(q_c)
    \end{bmatrix}.\]
    Here, $M_c(q_c)$ is the inertia matrix of the virtual instrument. In this instance we use virtual mass/inertia rather than using inerters, so that the behaviour of the virtual instrument is more similar to a real instrument acting as a constrained rigid body.

    The spring-dampers of Figure \ref{fig:VirtualInstrument} exemplify a more complex operation space than the standard Cartesian one.
    Consider the topmost spring-damper. The corresponding operation space coordinate is the component of the distance between the base of the real and virtual instruments \emph{in the x direction of the instrument's coordinate frame}.
    By using exactly 3 springs to constrain the 3 degrees of freedom of the instrument, we remain amenable to the parameter tuning technique of Section \ref{sec:Tuning} which requires that the number of spring-dampers be equal to the degrees of freedom of the robot.

    \subsection{Discussion and Comparison}

\begin{figure}[htbp]
        \centering
        \begin{subfigure}{0.4\columnwidth}
            \centering
            \vspace{8mm}
            \includegraphics{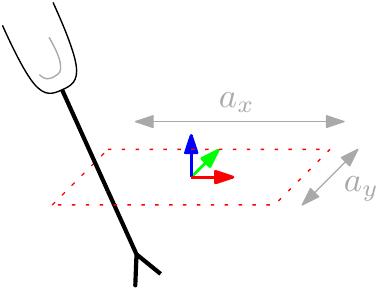}
            \caption{Rectangular constraint}
            \label{fig:Rect_a}
        \end{subfigure}%
        \begin{subfigure}{0.6\columnwidth}
            \centering
            \includegraphics{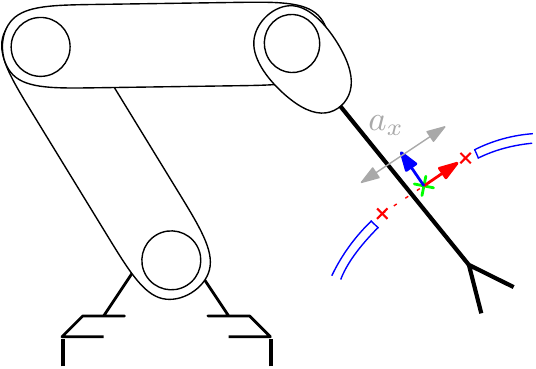}
            \caption{Section view}
            \label{fig:Rect_b}
        \end{subfigure}
        \caption{Rectangular relaxation of the RCM constraint, for springs with deadzone width $a_x$ and $a_y$.}
        \label{fig:Rect}
    \end{figure}

    The prismatic extension can easily be modified to allow a relaxation of the port constraint.
    In 3D, this can be achieved by rotating the springs on the slider such that they are aligned with the tangent-plane and normal vector of the patient's body, and replacing in-plane springs with a pair of piecewise-linear springs.
    These piecewise linear springs could have a low/zero stiffness region---a `deadzone'---with the effect of relaxing the point constraint of the RCM to a rectangular constraint in the specified plane. This is illustrated in Figure \ref{fig:Rect}. Other geometries (e.g. a disk constraint rather than a rectangle) correspond simply to an alternate choice of coordinate.

    Similar relaxations can be explored for the case of the virtual instrument. This second approach also offers a less aggressive use of the motor torques at the joints, reducing bandwidth requirements.
    The placement of the slider springs in the prismatic extension mechanism causes a loss in performance as the end-effector approaches the port.
    The `lever arm' between the end effector and RCM becomes small, thus the rotational stiffness imparted by the virtual mechanism to the instrument decreases.
    Conversely, as the instrument moves far away from the port (for example, when retracted from the patient), the lever arm and apparent rotational stiffness increase.
    This can result in a demanding torque signal for the motors, with higher bandwidth.
    The virtual instrument mechanism addresses these problems by placing the springs in the reference frame of the instrument, such that the instrument's rotational stiffness does not vary with the robot's pose.

    \begin{remark}
        Recent approaches in robotic surgery use hierarchical/constraint based techniques, such as \cite{Dietrich2020, Su2020a, Minelli2022}, which we briefly describe.
        If $z(q)=0$ is a desired constraint, then $\ddot{z} = J(q)\ddot q + \dot J(q) \dot q$,
        where the Jacobian $J(q)=\frac{\partial z}{\partial q}$.
        The input is chosen so that $\ddot q$ guarantees $J(q)\ddot q + \dot J(q)\dot q = 0$.
        The remaining degrees of freedom at the input are then used for lower priority task (nullspace projection method).
        However, in surgical applications there is no clear hierarchy between end-effector control and RCM control; failure in either poses sever consequences. The passivity of the nullspace projection method is also questionable, which potentially reduces the robustness of the controller, especially during interaction. In contrast, our virtual mechanism approach for RCM and end-effector control guarantees passivity, and performance is determined by the choice of the control parameters.
        \hfill $\lrcorner$
    \end{remark}

    \subsection{Virtual Mechanism Implementation}

\begin{figure}[b]
        \centering
        \includegraphics[width=0.95\columnwidth]{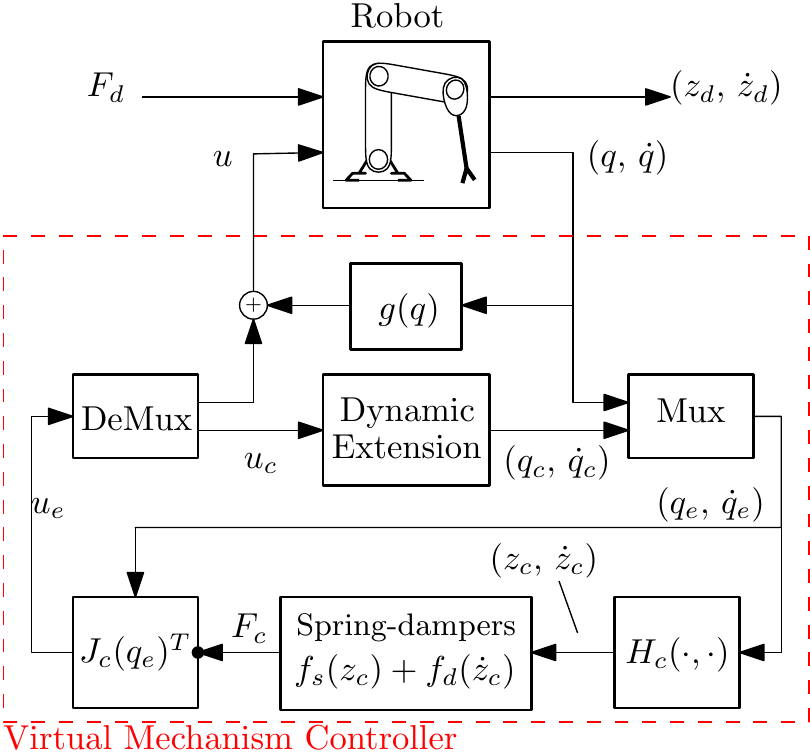}
        \caption{Virtual mechanism controller implementation.}
        \label{fig:MediumBlockDiagram}
    \end{figure}

    To implement these controllers, we must determine the corresponding joint-torques that represent the effect of the virtual mechanism. These are calculated using the operation-space Jacobian, as described in Section \ref{sec:ControllerStructure}.
    The implementation is illustrated by the  block diagram in Figure \ref{fig:MediumBlockDiagram}.
    Forces/positions $F_d$ and $z_d$ are exogenous inputs and outputs which we will use in Section \ref{sec:Tuning}.

    The dynamic extension block is an Euler-Lagrange system modelling virtual joints and links of the prismatic extension and of the virtual instrument, using the generalised coordinate $q_c$ .
    The extended state $q_e \triangleq \begin{bmatrix}q^T & q_c^T \end{bmatrix}^T$ is the concatenation of $q_c$ with the robot state, as indicated by the multiplex (Mux) block.
    The input $u$ is split between the robot's actuators and dynamic extension's actuators by de-multiplexing (DeMux). The actuator commands are sent to the robot, and the dynamic extension's state is updated in the controller, according to the dynamic equation in generalised coordinates $q_c$
    \begin{equation}
        \label{eq:dynamicExtension}
        M_c(q_c)\ddot q_c + C_c(q, \dot q)\dot q = u_c,
    \end{equation}
    where $M_c$ and $C_c$ are the inertia and Coriolis matrices of the dynamic extension.
    This implementation assumes the block-diagonal structure of the extended inertia matrix, with off diagonal blocks being zero, as described in Section \ref{sec:VirtualMechanismsForSurgery}.
    Mechanisms without this separation of inertia are possible, but the required additional inputs for the extra inertial forces would have to be provided to the both the robot and dynamic extension.

    If we define a forward kinematics function $h_c(q_e)$ which returns the extension of the virtual mechanism's springs $z_c$, then its partial derivative gives the Jacobian
    \[J_c(q_e)  \triangleq  \dfrac{\partial h_c(q)}{\partial q},\]
    and together they define $H_c$:
    \[(q_e, \dot q_e) \rightarrow (h_c(q_e), J_c(q_e) \dot q_e) = (z_c, \dot z_c),\]
    $J_c$ is the interconnecting Jacobian of Figure \ref{fig:PortHamiltonian} with extra columns to relate motion/forces from/to the generalised coordinates of the dynamic extension.

    The controller force, $F_c$, is the combined force of all springs and dampers in operation space.
    If the springs/dampers are linear, then the general formulation $f_s(z_c) + f_d(\dot z_c)$ can be replaced with
    \begin{equation}
        \label{eq:Fc}
        F_c=\underbrace{\begin{bmatrix}\diag(k)& \diag(c)\end{bmatrix}}_{=K}\begin{bmatrix}z_c \\ \dot z_c\end{bmatrix},
    \end{equation}
    where $k$ is the vector of spring-stiffnesses and $c$ of damping coefficients.
    $K$ represents a linear state-feedback matrix in operation space $z_c$.
    Note that the operation space coordinates are designed such that $z_c = 0$ is the desired equilibrium, by including the appropriate constant offset in the transform $H_c$.

    \section{Parameter tuning} \label{sec:Tuning}

    \subsection{Controller gain design via linear matrix inequalities}

    The second step in controller synthesis is the selection of controller parameters/gains.
    Using the technique of \cite{Larby2022} we design the stiffness and damping parameters of the virtual mechanism controller to (i) guarantee passivity at any port $(F_e, \dot z_e)$ of the robot-virtual mechanism controller closed-loop as shown in Figure \ref{fig:PortHamiltonian}; and to (ii) maximise performance relative to the exogenous input/output pair $(F_d, (z_d, \dot z_d))$, as shown in Figure \ref{fig:MediumBlockDiagram}. The latter will be established locally around specific poses $q_i$.
    The key theorem is summarized as follows.

    \begin{thm}
        \label{thm:main_result}
        Consider the closed-loop system of Figure \ref{fig:MediumBlockDiagram}.
        Let $q_i$ be a selection of poses for $i \in \{1, \dots, N\}$.
        Let $z_c$ be a vector of dimension $n$, where $n$ is the combined number of degrees of freedom of the robot and dynamic extension.

        Then, for any selected gain $\gamma>0$ and scaling matrices $W_1$ and $W_2$ of suitable dimension, the feasibility of the LMI condition of Theorem 2 in \cite{Larby2022} in the variables $L$ and $Q$ guarantees that:
        \begin{itemize}
            \item $[\diag(k), \, \diag(c)] = LQ^{-1}$ in \eqref{eq:Fc}
            \item the closed-loop is passive at any port $(F_e, \dot z_e)$
            \item the closed-loop satisfies the local $\mathcal{L}_2$ gain
            $$\left\| \begin{bmatrix} W_1 z_d \\ W_2 \dot{z_d} \end{bmatrix} \right\|_2 \leq \gamma \|F_d\|_2$$
            for an initial condition sufficiently close to any of $q_i$, for any sufficiently small $F_d$.
        \end{itemize}
    \end{thm}
    \vspace{2mm}

    For more details on Theorem 2, the specific formulation of the linear matrix inequalities, and a complete proof, we refer the reader to \cite{Larby2022}.  Here, for completeness, we observe that the LMI formulation is always feasible for the minimal case given by $N=1$ and $\gamma$ sufficiently large, provided that $q_i$ does not correspond to a point of singularity for the robot-virtual-mechanism controller

    \begin{figure}[htbp]
        \centering
        \includegraphics[width=\columnwidth]{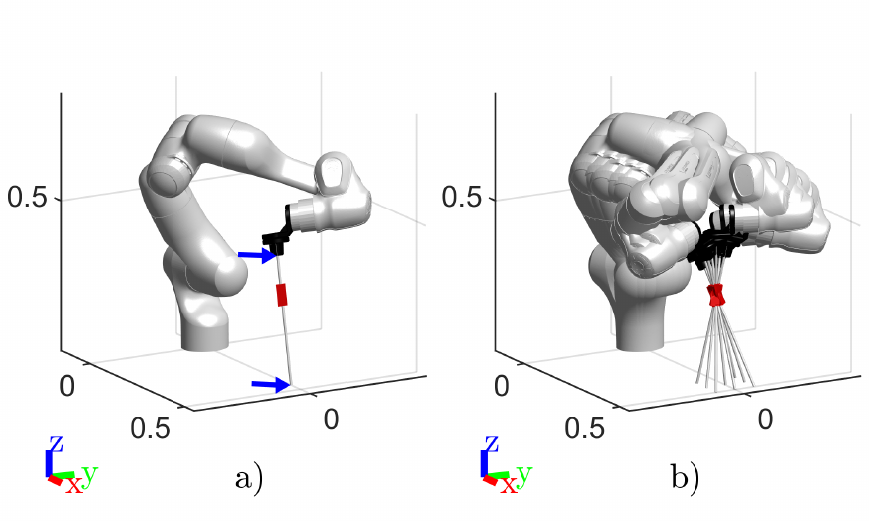}
        \vspace{-8mm}
        \caption{Franka-Emika Robot URDF model with prismatic extension a) in centre pose with exogenous input/output locations marked by the blue arrows, and b) in 9 poses on the $3\times3$ grid (right).}
        \label{fig:robotPoses}
    \end{figure}

    \subsection{Task specification}

    For the formulation of the linear matrix inequalities of Theorem 2, we require:
    \begin{enumerate}
        \item A robot dynamic model
        \item Exogenous inputs and outputs, $F_d$, $z_d$ and $\dot z_d$
        \item Controller operation space $z_c$, which determines the placement of the $n$ spring-dampers
        \item A set of poses $q_i$ for $i \in \{1, \dots, N\}$, around which to linearise and optimize performance
        \item Weighting terms $W_1$ and $W_2$, and $\gamma$
    \end{enumerate}

    We use a URDF model of the Franka-Emika robot arm modified to feature a surgical instrument,
    as shown in \ref{fig:robotPoses}.
    We specify the inertance of the prismatic extension $m_c = \SI{0.1}{kg}$, and the mass/inertia of the virtual instrument to be $m_i = \SI{0.1}{kg}$,  $ I_i = \diag(.01, .01, .01) \SI{}{kg\cdot m^2}$.
    These values are chosen so as to be small enough to not add lots of extra inertia, but not so small as to greatly increase the number of required simulation timesteps.

    For the relevant exogenous inputs/outputs, we will first consider the input given by a disturbance force acting on the end-effector, taking as output its position/velocity.
    We will also consider a force applied at the base of the instrument (input) and the associated position/velocity output.
    These are combined into $6\times1$ vectors of forces and displacements, and velocities, $F_d$, $z_d$ and $\dot z_d$.
    These inputs/outputs capture the requirements that there should be a high impedance from force to motion at the end-effector, and small motion at the trocar.
    We will take $\gamma = 1$, $W_2=0$, and vary $W_1$.

    Based on the synthesis in \cite{Larby2022}, the number of spring-dampers must equal the degrees of freedom of the closed-loop system (DOF of robot + DOF of virtual mechanism = number of spring-dampers).
    The Franka-Emika robot has $7$ degrees of freedom;
    the prismatic extension adds 1 further joint/DOF and 6 spring/dampers;
    the virtual instrument adds 3 further joints/DOF and 8 spring/dampers.
    Therefore, both controllers require 2 additional spring dampers to apply the synthesis.
    We choose to use jointspace spring-dampers on J2 and J4\footnote{These refer to the second and fourth joints, counting from the base.}, with a resting position at $\pi/6$ and $-4\pi/6$ respectively. These particular choices of J2/J4 angles are to keep the arm upright and with J4 folded, but away from singularities. The operation space variables (i.e. spring damper placement) for each mechanism are summarised in Tables \ref{tab:PE} and \ref{tab:VIzc}.

    The set of joint poses $q_i$ about which the robots dynamics will be linearised, and performance optimized, are chosen as follows:
    9 poses passing through an RCM at $(0.4, 0, 0.35)$ to place the end-effector on an evenly spaced $3\times3$ grid in the xy plane, with corners at $(0.4, -0.05, 0.15)$ and $(0.5, 0.05, 0.15)$, and with J2 and J4 equal to $\pi/6$ and $4\pi/6$ respectively.
    The model robot and poses are shown in Figure \ref{fig:robotPoses}.
    Care must be taken when choosing poses to avoid singularities. If the controller Jacobian is near singular, then the gains required to achieve a particular performance at that pose will be large.

    Synthesis results for the prismatic extension and for the virtual instrument are shown within Tables \ref{tab:PE} and \ref{tab:VIzc}, respectively.
    Note that the stiffness/damping parameters for J2, J4 and the instrument-aligned spring-damper for the virtual instrument mechanism are small.
    The synthesis ensures that they are greater than zero such that the linearisation is stable, but they are not necessary for minimising the gain between the chosen $F_d$ and $z_d$.
    Increasing $W_1$ has the desired effect of increasing the overall gains.

    \newcommand{\specialcell}[2][c]{%
        \begin{tabular}[#1]{@{}l@{}}#2\end{tabular}}

    \begin{table}[htbp]
        \hfill
        \begin{tabular}{l!{\vrule width 1pt}l}
            $z_c$  & Description \\
            \hline
            1:3 & \specialcell{Vector from reference \\ to end effector position,\\ in  world frame} \\
            \hline
            4 & J2 angle: $q_2 - \pi/6$ \\
            \hline
            5 & J4 angle: $q_4 + 4\pi/6$ \\
            \hline
            6:8 & \specialcell{Vector from RCM to\\ virtual prismatic ext-\\ension  position, in \\ world frame} \\
        \end{tabular}%
        \hfill
        \begin{tabular}{r|rr}
            &\multicolumn{2}{c}{$W_1=20I$}  \\
            &$k$ & $c$ \\
            \hline
             $1$ &  $2500$ &  $401$ \\  $2$ &  $3180$ &  $572$ \\  $3$ &  $1940$ &  $277$ \\  $4$ &  $0.131$ &  $0.56$ \\  $5$ &  $0.148$ &  $1.2$ \\  $6$ &  $3470$ &  $810$ \\  $7$ &  $4230$ &  $984$ \\  $8$ &  $1930$ &  $359$
        \end{tabular} \vspace{2mm}
        \captionsetup{width=1\columnwidth}
        \caption{Spring-damper placement for the prismatic extension (left). Synthesized stiffness and damping parameters, $k$ and $c$ (right).}
        \label{tab:PE}
    \end{table}

    \begin{table}[htbp]
        \begin{tabular}{l!{\vrule width 1pt}l}
            $z_c$  & Description \\
            \hline
            1:3 & \specialcell{Vector from reference to end effector position, in \\  world frame} \\
            \hline
            4 & J2 angle: $q_2 - \pi/6$ \\
            \hline
            5 & J4 angle: $q_4 + 4\pi/6$ \\
            \hline
            6:7 & \specialcell{x/y component of vector from virtual end-effector to \\ end-effector, in instrument frame} \\
            \hline
            8 & \specialcell{z component of vector from virtual instrument centre \\ to instrument centre, in instrument frame} \\
            \hline
            9:10 & \specialcell{x/y component of vector from virtual instrument base to \\ instrument base, in instrument frame}
            \vspace{2mm}
        \end{tabular}

        \begin{tabular}{r|rr|rr}
            &\multicolumn{2}{c|}{$W_1=20I$} & \multicolumn{2}{c}{$W_1=100I$}  \\
            & $k$ & $c$ & $k$ & $c$  \\
            \hline
             $1$ &  $1420$ &  $250$ &  $7720$ &  $557$ \\  $2$ &  $1780$ &  $349$ &  $9930$ &  $779$ \\  $3$ &  $1150$ &  $172$ &  $7230$ &  $432$ \\  $4$ &  $0.152$ &  $0.782$ &  $0.149$ &  $0.809$ \\  $5$ &  $0.156$ &  $1.25$ &  $0.157$ &  $1.41$ \\  $6$ &  $428$ &  $45.9$ &  $3000$ &  $119$ \\  $7$ &  $1470$ &  $306$ &  $10500$ &  $817$ \\  $8$ &  $0.0515$ &  $0.0913$ &  $0.0513$ &  $0.0871$ \\  $9$ &  $884$ &  $89.8$ &  $5060$ &  $197$ \\  $10$ &  $2860$ &  $727$ &  $15200$ &  $1570$
        \end{tabular} \vspace{2mm}
        \centering
        \captionsetup{width=1\columnwidth}
        \caption{Spring-damper placement for the virtual instrument (top). Synthesized stiffness and damping parameters, $k$ and $c$, for two different weight terms $W_1$ (bottom).}
        \label{tab:VIzc}
    \end{table}

    \vspace{-2mm}
    \section{Results} \label{sec:Results}
    \vspace{-2mm}

\begin{figure*}[htbp]
        \centering
        \includegraphics[]{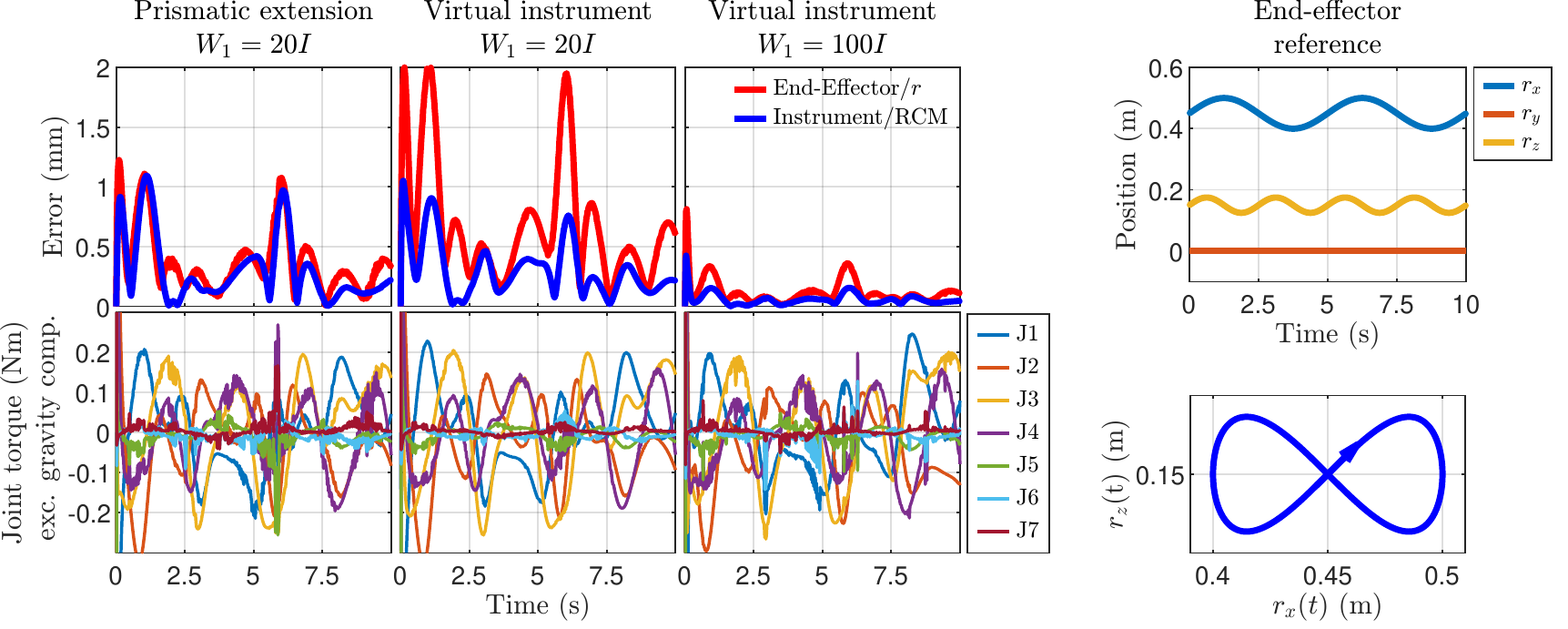}
        \vspace{-5mm}
        \caption{Simulation results.}
        \label{fig:Results}
    \end{figure*}

    We performed simulations for each set of gains, using the URDF model of the robot following the desired end-effector trajectory
    \begin{equation}
        \label{eq:eightcurve}
        r(t) = \begin{bmatrix}
            0.45 + 0.05 \sin(2\pi/5) \\
            0 \\
            0.15 + 0.025 \sin(2\pi/10)
        \end{bmatrix},
    \end{equation}
    represented in Figure \ref{fig:Results}.

    Also shown in Figure \ref{fig:Results} are the results of the simulation, where the absolute error between the end-effector tip is plotted alongside the minimum distance between the axis of the instrument shaft and the remote centre of motion. Joint torques are shown without gravity compensation, which would otherwise dominate the plot.

    The prismatic extension and virtual instrument perform similarly, when using gains synthesized with the same weighting and exogenous inputs/outputs. The increased gains as a result of setting $W_1$ to $100$ reduce the error further, at the cost of a higher bandwidth controller.

\begin{figure}[htbp]
        \centering
        \includegraphics[]{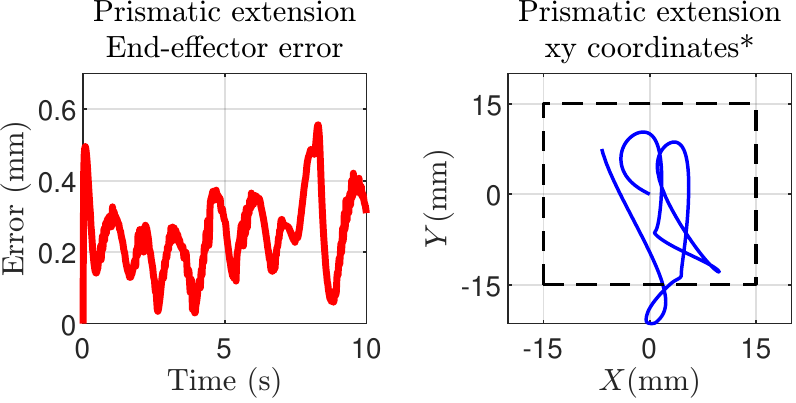}
        \vspace{-2mm}
        \caption{Simulation results for prismatic extension with relaxed RCM constraint. *relative to the RCM.}
        \label{fig:piecewise}
    \end{figure}

    We also present results from a simulation using the rectangular constraint of Figure \ref{fig:Rect}.
    The X and Y Cartesian springs of the prismatic extension (i.e. on the 6th and 7th coordinates of $z_c$) are replaced with piecewise springs, having zero stiffness regions of width/depth $a_x = a_y = \SI{30}{mm}$, with the outer stiffness taken from Table \ref{tab:PE}.
    The corresponding dampers have their damping coefficients reduced to 5\% of the values in Table \ref{tab:PE}.
    All other spring-dampers are unchanged.
    Figure \ref{fig:piecewise} shows an improvement in the end-effector performance, as compared with the first column of Figure \ref{fig:Results}, due to the relaxation of the RCM constraint.
    Due to space constraints, we will address the challenges of parameter synthesis for piecewise springs in future work.

    \vspace{-3mm}
    \section{Conclusions}
    \vspace{-2mm}

    In this paper we have addressed the problem of controller design for robotic minimally invasive surgery by means of a virtual mechanism design, and presented two solutions. We have also demonstrated an optimization based approach for tuning the parameters of these two controllers. Overall, this result in controllers that guarantees passivity and local $\mathcal{L}_2$ gain $\gamma$ between selected inputs/outputs. Results are illustrated in simulations.

    In the last section we have also discussed a simple relaxation of the remote centre of motion constraint  to rectangular constraints, using piecewise springs, as shown in Figure \ref{fig:Rect} and in Figure 9.
    For reasons of space we could not provide details on the synthesis of piecewise linear springs. These will be left to future publications.
    This can be approached via the linear-matrix-inequality technique by having more variables  $(L, Q)$ corresponding to multiple regions.
    Each pose $q_i$, $i\in{1, \dots, N}$ belongs to a region, and each region features differing weighting $W_1$, $W_2$.
    Additional equality constraints are used between $(L, Q)$ from different regions to keep constant parameters which do not vary between particular pairs of regions.

    While the mechanisms in this paper use precisely $n$ spring-dampers, such that we can apply the synthesis technique of \cite{Larby2022}, the virtual mechanism approach allows arbitrarily many springs and dampers, lending greater flexibility and possibly better performance.
    One potential approach to synthesis for a number of spring-dampers greater than $n$ involves updating $n$ parameters at a time (using the linear matrix inequality synthesis technique) while keeping fixed the other parameters.
    At each iteration, a different set of $n$ spring-damper parameters is fixed, until the whole set of parameters converges to a stable solution.
    Work is needed to determine the feasibility of this approach.

    Another possible approach to virtual mechanism parameter tuning is the use of gradient-based approach to minimize a chosen cost function over a trajectory or set of trajectories.
    This approach is less constrained than the linear matrix inequality approach, allowing optimization over any number of parameters, type of controller parameter (e.g. inertances, masses, geometries), and non-linear components.
    The challenge lies in the choice of appropriate cost functions and in the optimization of the computation.

    \vspace{-3mm}
    \bibliography{../../../zotero/library}             %

\end{document}